\documentclass[review]{elsarticle}

\usepackage{lineno,hyperref}
\modulolinenumbers[5]

\journal{Journal of \LaTeX\ Templates}

\usepackage{dblfloatfix}
\usepackage{array,multirow,graphicx}
\usepackage{multirow}
\usepackage{diagbox}
\usepackage{gensymb}
\usepackage[table]{xcolor}
\usepackage{amsmath}
\usepackage{amssymb}
\usepackage{amstext}
\usepackage{amsthm}
\usepackage{mathtools}

\usepackage{algorithmic}

\usepackage{epsfig}
\usepackage{calc}

\usepackage{tikz}

\usepackage{pifont}

\DeclarePairedDelimiterX{\norm}[1]{\lVert}{\rVert}{#1}

\newcommand{\benjamin}[1]{\textcolor{black}{#1}}
\newcommand{\marius}[1]{\textcolor{black}{#1}}

\begin{document}

\begin{frontmatter}

\title{A comparative study on Optical Flow \\ for Facial Expression Analysis}





\author[label1]{B.Allaert\corref{cor1}}
\ead{benjamin.allaert@imt-nord-europe.fr}
\author[label3]{IR.Ward}
\ead{isaac.ward@uwa.edu.au}
\author[label2]{IM.Bilasco}
\ead{marius.bilasco@univ-lille.fr}
\author[label2]{C.Djeraba}
\ead{chabane.djeraba@univ-lille.fr}
\author[label3]{M.Bennamoun}
\ead{mohammed.bennamoun@uwa.edu.au}
\cortext[cor1]{Corresponding author}
\address[label1]{IMT Nord Europe, Institut Mines-Télécom, Univ. Lille, Centre for Digital Systems, F-59000 Lille, France.}
\address[label2]{Centre de Recherche en Informatique Signal et Automatique de Lille, Univ. Lille, CNRS, Centrale Lille, UMR 9189 - CRIStAL -, F-59000 Lille, France.}
\address[label3]{University of Western Australia (M002), 35 Stirling Highway, 6009 Perth, Australia.}

\begin{abstract}
Optical flow techniques are becoming increasingly performant and robust when estimating motion in a scene, but their performance has yet to be proven in the area of facial expression recognition. In this work, a variety of optical flow approaches are evaluated across multiple facial expression datasets, so as to provide a consistent performance evaluation. The aim of this work is not to propose a new expression recognition technique, but to understand better the adequacy of existing state-of-the art optical flow for encoding facial motion in the context of facial expression recognition. Our evaluations highlight the fact that motion approximation methods used to overcome motion discontinuities have a significant impact when optical flows are used to characterize facial expressions.
\end{abstract}


\end{frontmatter}


\section{Introduction}
\label{sec:introduction}

Building a system that is capable of automatically recognizing the emotional state of a person from their facial expressions has been a burgeoning topic in computer vision in the recent years. Automating the analysis of facial expressions, from videos, is highly beneficial in a range of diverse applications, including security, medicine, and human-machine interaction. For instance, the analysis of the emotional state of a patient, based on their facial expressions, can help to estimate the quality of the provided care, and to monitor the ongoing patient-doctor relationship.

The use of facial expression information increases proportionally with our need to automate the process of extracting behavior and cognitive-related information (expressions, intentions and predictions). Although many advances have been achieved in this area, the recent approaches do not yet achieve satisfactory results when deployed in real-world situations (e.g., in transportation and retail stores).

Considering the nature of the features used to characterize facial expressions, the majority of the existing approaches are based on texture or geometry \cite{lv20193d, tran2021micro}. Yet, the analysis of the facial movement through optical flow seems to offer a promising avenue of research for expression analysis. It is mainly used for its ability to characterize both intense and subtle movements \cite{allaert2018advanced}, as well as being able to correct head pose variations \cite{yang2017dense}, micro-expression spotting \cite{wang2017main} , or to deal with facial occlusions \cite{poux2020facial,poux2021dynamic}.

\begin{figure}[!h]
\centering
\includegraphics[width=\columnwidth]{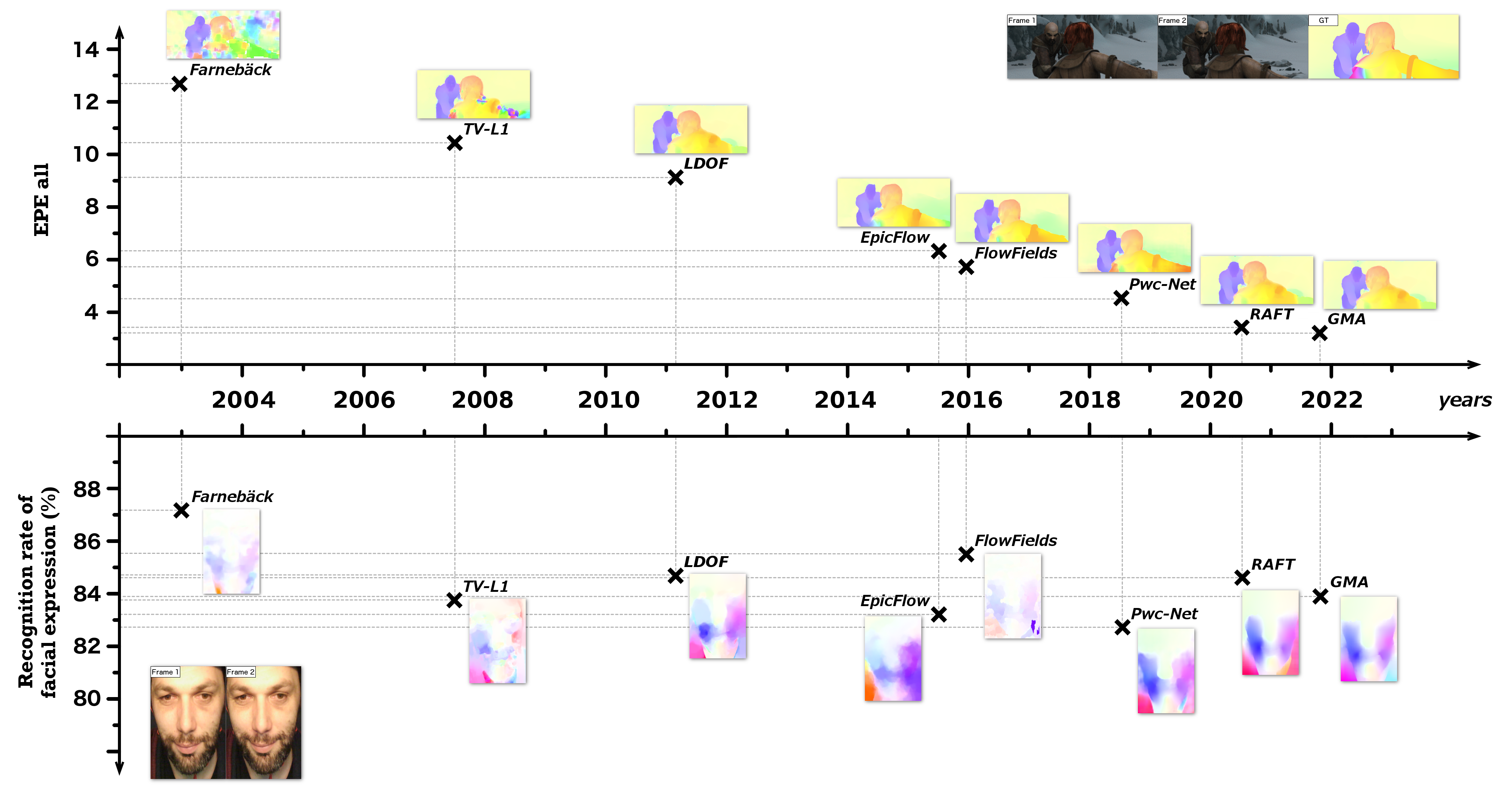}
\caption{\benjamin{Comparison of the performance of several optical flow approaches on the MPI-Sintel, a generic synthetic movie dataset (top) for optical flow analysis, and on a set of facial expression datasets (bottom). Such datasets are used due to the lack of optical flow ground-truth data for facial expression analysis. Although the performance tends to be conclusive on MPI-Sintel, this is not the case for facial expression analysis, where a basic approach such as Farnebäck gives the best performance.}}
\label{fig:intro}
\end{figure}

A number of methodological innovations have progressively been introduced to improve the performance of dense optical flow techniques on datasets, such as MPI-Sintel \cite{Butler:ECCV:2012}, as illustrated in the top of Figure \ref{fig:intro}. However, several authors suggest that the use of the \textit{recent} optical flow approaches tends to \textit{reduce} the system performance in fields such as human action recognition \cite{wang2013dense,martin2019optimal} or facial expression recognition \cite{snape2015face} in comparison with traditional 
optical flow techniques --- such as the one proposed by Farnebäck \cite{farneback2003two} (reflected in the bottom of Figure \ref{fig:intro}). Nevertheless, no clear protocol of comparison has yet been proposed.

To understand this paradox, this work investigates the impact of the most recent dense optical flow approaches on the performance of facial expression recognition. More specifically, this study is the first attempt to address the question: "Which optical flow approach should one use to analyze facial motion?". 

The paper is organised as follows: we introduce, in Section \ref{work}, the challenges of detecting facial motion (especially motion discontinuities), and we briefly describe the main characteristics of the major optical flow techniques that are proposed in the literature. In Section \ref{baseline}, we introduce the datasets that we used to compare our selected optical flow approaches and define their performance criteria. We then evaluate the capacity of our selected optical flow approaches to accurately detect facial movements, by combining them with different handcrafted and learning-based approaches on a variety of other facial expression datasets in Section \ref{solo}. In Section \ref{fusion}, we analyze the use of distinctive features of different optical flow approaches to artificially augment the training dataset. To conclude, we summarize our results and discuss future perspectives in Section \ref{conclu}.

\section{Scope and background}
\label{work}

This section highlights the main objectives of the paper, lists the major developments of dense optical flow approaches and gives an overview of the optical flow approaches that we selected for this analysis and their characteristics.

\subsection{Scope of the paper}

The paper is essentially focused on evaluating the adequacy of existing optical flow techniques for encoding the facial movement in view of expression recognition. Although, nowdays, deep learning architectures that encodes the spatio-temporal information for expression recognition overtake handcrafted feature extraction, we think it is still valuable to discuss existing optical flow techniques (handcrafted or based on deep learning). Employing existing optical flow techniques can be, in a way, seen as a transfer learning approach, reducing the training process and the overall architecture footprint. In the effort of deploying facial expression analysis tool on restricted resources devices it is important to be able to dispose of lightweight architectures that takes advantage of state-of-the art descriptors (handcrafted or learnt).

The unique characteristics of facial movement implies that some motion discontinuities tend to provide information about an expression \cite{cao2015real}. Therefore, the need to devise dense optical flow approaches to address motion discontinuities, while ensuring a rapid computation time, is both an important requirement and a significant challenge. Consequently, it is important to study how optical flow techniques deal with motion discontinuities, while being immune to noise propagation in the neighboring regions.

Although optical flow approaches are becoming more and more robust on datasets such as MPI-Sintel (a dataset for scene analysis), it is important to consider the performance of these approaches for facial expression analysis. As stated, the challenges proposed by the data of MPI-Sintel do not always reflect the problems that can be observed on a face. Indeed, MPI-Sintel is a synthetic dataset generated through 3D rendering that contains optical flow ground-truth extracted from a generic synthetic movie. No specific information about facial data is provided. In this context, relying on the results obtained by optical flow approaches on MPI-Sintel may not be relevant in identifying the best optical flow approaches to characterize facial motion. 

In this work, we have voluntarily chosen to study optical flows on datasets specific to macro expressions. These types of expressions allow us to focus on movements of higher intensity, and to better compare our studies with reference datasets of optical flows such as MPI-Sintel. Moreover, we have chosen only datasets where the movement of facial expressions is not noisy (as it can be the case on the in-the-wild datasets). It is indeed important to study how optical flow approaches manage to encode micro-expressions. However, we have voluntarily not studied this issue in the paper, to avoid the presentation of two different evaluation protocols for macro and micro expression (labeling, classification). We focused in this paper on evaluation protocol of macro expression.

In this paper we provide three key contributions. \textbf{First}, we study the ability of different optical flow approaches in characterizing facial expressions. The key processing steps of each approach are analyzed in order to identify those which have a tendency to improve or reduce performance (Section \ref{solo}). \textbf{Second}, we investigate whether several optical flow approaches can be used collaboratively to characterize facial movements, in place of using a single, highly performant approach. Put more explicitly, we answer the following question: "Can optical flow approaches be used in a data augmentation process in the context of deep learning architectures?" To answer this, we analyze several optical flow approaches and their characteristics --- taking particular note of how they handle motion discontinuities (Section \ref{fusion}). \textbf{Finally}, in order to benchmark and compare the results of our work, we propose a new evaluation baseline for evaluating the performances obtained by using \textit{several} optical flows on various facial expression datasets, by comparing different handcrafted and deep-learning-based algorithms. In order to ensure that our experiments are reproducible, all the data are made available online on demand. 

\subsection{Background of the optical flow techniques}
\label{ssec:background}

Optical flows are relatively sensitive in the presence of some factors such as occlusions, light changes or out-of-plane movements. All these factors lead to the appearance of false movements, which result in motion discontinuities. Training benchmarks such as MPI-Sintel \cite{Butler:ECCV:2012} have been proposed in order to address these problems. In response, many optical flow approaches have been proposed. Some of these approaches are distinguished by their originality in regards to how they implement some key processing steps, including the matching, filtering, interpolation and optimization. 

Dense optical flow approaches are based on several strong assumptions about the properties of motion flow. Among these assumptions, we can note : the structure (preservation of the shapes and edges of the objects in the scene) and motion adeptness (specific to each object) \cite{wedel2009structure}, the approximation by motion models \cite{chen2013large} and the local regularity \cite{horn1981determining}. The local regularity hypothesis is generally applied through a joint energy-based regularization that evaluates the coherence and smoothness of the motion variations. The major drawback is that rapid minimization techniques generally rely on local linearization and can therefore only adapt the motion field very locally. Therefore, these methods must use pyramidal approaches to deal with large displacements \cite{farneback2003two}. In practice, this fails in cases where the motion determined at the lower resolution level is not consistent with the motion at the higher resolution level.

To overcome this problem, rapid approaches allow to efficiently perform a global search to find the best match on the image at the top level \cite{he2012computing, weinzaepfelICCV2013}. However, since there is no regularization, neighboring fields usually contain many outliers that are difficult to identify. In addition, even if outliers can be identified, they leave empty motion field gaps that need to be filled.

In order to reduce the number of outliers in motion propagation, some approaches rely on sparse descriptor matching \cite{wedel2009improved, brox2011large, weinzaepfelICCV2013}. This consists in relying only on regions where the movement is highly consistent. However, due to the scarcity of such regions, these approaches tend to induce more outliers than approximation approaches based on neighboring regions. The differences between the estimated motion and the actual motion can be relatively large since a motion for which no match is found cannot be taken into account. Despite these difficulties, approaches based on sparse descriptors or on the propagation of motion from neighboring regions have become increasingly popular in recent years as an initial step in large displacement optical flow algorithms \cite{revaud2015epicflow,sun2018pwc}.

However, while there are descriptor-matching approaches such as Deep Matching \cite{weinzaepfelICCV2013} that are adapted to the optical flow, dense initialization is usually simply based on the approximation of motion based on the neighboring regions - which is sub-optimal. The intention behind the motion approximation of the neighboring regions is to find the closest match visually, which is often not identical to the expected motion. An important difference is that motion in neighboring regions is known to be very noisy with respect to the shift of the neighboring pixels, whereas the optical flow is generally locally smooth and sometimes abrupt.

Recent approaches tend to solve these problems not by explicit regularization or smoothing (such as median filtering), but by proposing adapted search strategies for finding most outliers while avoiding to propagate them \cite{bailer2015flow}. These approaches contain far fewer outliers than those based on the approximation of the motion of neighboring regions with respect to optical flow estimation.

\benjamin{Usually, the optimisation objective defines a trade-off between a data term that encourages the alignment of visually similar image regions and a regularisation term that imposes priors on the plausibility of motion. This approach proved to be very successful, but further progress has been challenging due to the difficulties of manually designing an optimisation objective that is robust to a variety of special cases. Recently, deep learning has emerged as a promising alternative to handcrafted methods. Current deep learning methods have achieved performance comparable to the best traditional methods while being significantly faster at inference. Among the recent approaches, the RAFT \cite{teed2020raft} architecture stands out from other architectures. Unlike traditional approaches, features and motion priorities are not created by hand but learned by the feature encoder and update operator, respectively. Recent extensions have been proposed to improve the performance of this system, such as adapting the upsampling method to exploit the fine details during training \cite{eldesokey2021normalized} or modelling self-similarities between two successive images to make the system more robust to occlusions \cite{jiang2021learning}.}

\subsection{Selected Optical Flow techniques}

To evaluate the impact of different optical flow approaches on the analysis of facial expressions, we have selected \benjamin{nine} approaches \marius{(listed chronologically bellow)} amongst an exhaustive panel of approaches that cover the different technical developments mentioned above. \marius{Currently, } 
\benjamin{335 optical flow methods have been evaluated and \marius{ranked over the last ten years} on the MPI-Sintel reference database. For each optical flow method selected below, we also mention their ranking.} The techniques which are of interest to this work are \textit{briefly} described here. For a quantitative comparison, please see Section \ref{solo}.

\textbf{Farnebäck's} method \cite{farneback2003two} \benjamin{(\marius{2003, unranked by MPI-Sintel)}
} embeds a translation motion model between neighborhoods of two consecutive images in a pyramidal decomposition. Polynomial expansion is employed to approximate pixel intensities in the neighborhood. The tracking begins at the lowest resolution level, and continues until convergence. The pyramid decomposition enables the algorithm to handle large pixel motions, including distances greater than the neighborhood size. 

\textbf{TV-L1} \cite{wedel2009improved} \benjamin{(\marius{2009}, rank 312 \marius{in MPI-Sintel})} is a particularly appealing formulation which is based on total variation (TV) regularization and the robust L1 norm in the data fidelity term. This formulation can preserve discontinuities in the flow field and thus offers an increased robustness against illumination changes, occlusions and noise.

\textbf{Ldof} \cite{brox2011large} \benjamin{(2011, rank 300 \marius{in MPI-Sintel})} estimates large movements in small structures by integrating the correspondences (from descriptor matching) into a variational approach. These correspondences are \textit{not} used in order to improve the accuracy of the approach; they are used as they support the coarse-to-fine warping strategy and avoid local minima.

\textbf{EpicFlow} \cite{revaud2015epicflow} \benjamin{(2015, rank 203 \marius{in MPI-Sintel})} relies on the deep matching algorithm integrated into the \textbf{DeepFlow} method \cite{weinzaepfelICCV2013} and interpolates a set of sparse matches in a dense manner to initiate the estimation. This approach preserves the edges so that they can be used in the interpolation of movement. The solution has proven its effectiveness in characterizing optical flows over multiple datasets, including MPI-Sintel \cite{Butler:ECCV:2012} and KITTI \cite{geiger2013kitti}.

\textbf{FlowFields} \cite{bailer2015flow} \benjamin{(2015, rank 171 \marius{in MPI-Sintel})} use a dense correspondence field technique that is \textit{much less} outlier prone. This method does not require explicit regularization or smoothing (such as in median filtering), but is instead a pure data-oriented search strategy which only finds most inliers, while effectively avoiding the outliers.

\textbf{PWC-net} \cite{sun2018pwc} \benjamin{(2018, rank 85)} is based on a compact CNN model which uses simple and well-established principles: pyramidal processing, warping, and the use of a cost volume. The particularity of this method is that the warping and the cost volume layers have no learnable parameters that can reduce the model size. As with most recent approaches, motion discontinuities are handled by post processing the optical flow using median filtering.

\benjamin{\textbf{RAFT} \cite{teed2020raft} \benjamin{(2020, rank 21 \marius{in MPI-Sintel})} maintains and updates a single fixed flow field at high resolution that overcomes several limitations of a coarse-to-fine cascade like the difficulty of recovering from errors at coarse resolutions and the tendency to miss small fast-moving objects. Second, the update operator of RAFT is recurrent and lightweight \marius{and} can be applied 100+ times during inference without divergence. Moreover, the update operator is based on a convolutional GRU that performs lookups on 4D multi-scale correlation volumes instead of only plain convolution or correlation layers used by others.}

\benjamin{\textbf{NCUP} \cite{eldesokey2021normalized} \benjamin{(2021, rank 11 \marius{in MPI-Sintel})} use a joint upsampling approaches within the coarse-to-fine optical flow CNN\marius{. In an end-to-end fashion, it allows} optical flow networks to exploit the fine details during training. This approach is based on a novel joint upsampling approach (inspired by PWC-net \cite{sun2018pwc}) that formulates the upsampling as a sparse problem and employs the RAFT \cite{teed2020raft} normalized convolutional neural networks to solve it.}

\benjamin{\textbf{GMA} \cite{jiang2021learning} \benjamin{(2021, rank 4 \marius{in MPI-Sintel})} uses a transformer-based approach to find long-range dependencies between pixels in the first image, and performs global aggregation on the corresponding motion features to overcome occlusions between two successive images. This method differs from other approaches that rely \marius{only} on CNNs to learn occlusions or require multiple images to reason about occlusions using temporal smoothing.}

\section{Datasets and performance criteria}
\label{baseline}

In this work, two primary sets of experiments are conducted: the evaluation of optical flow approaches (Section~\ref{solo}), and the augmentation of the training data through the use of optical flows (Section~\ref{fusion}). The datasets used and the performance criteria in each case are outlined in this section.

\subsection{Datasets}

There is no dataset which offers a ground-truth to accurately compare the performance of optical flow approaches against the task of characterizing facial movements. Thus, we first propose a baseline based on a set of facial expression datasets which contain different expression intensities. For the purpose of our work, it is necessary to analyze temporal sequences, and hence image datasets such as JAFFE, RaFD or AffectNet (where expressions are only represented by one image or several images that are not temporally successive) are not considered. As the main aim of this study is to evaluate the capacity of the optical flow approaches in characterizing facial movement, we select data acquired in controlled conditions, where only the movement related specifically to the facial expression is present. Datasets such as RECOLA and so forth, which contain numerous pose variations, occlusions and light changes are thus omitted from this study, as the biases induced by these challenges interfere with the native capacity of the optical flow approaches in characterizing facial movement. A standardization step would reduce these biases, but there is no guarantee that this will not have an impact on the quality of the resulting optical flows.

\begin{figure}[!h]
\centering
\includegraphics[width=\columnwidth]{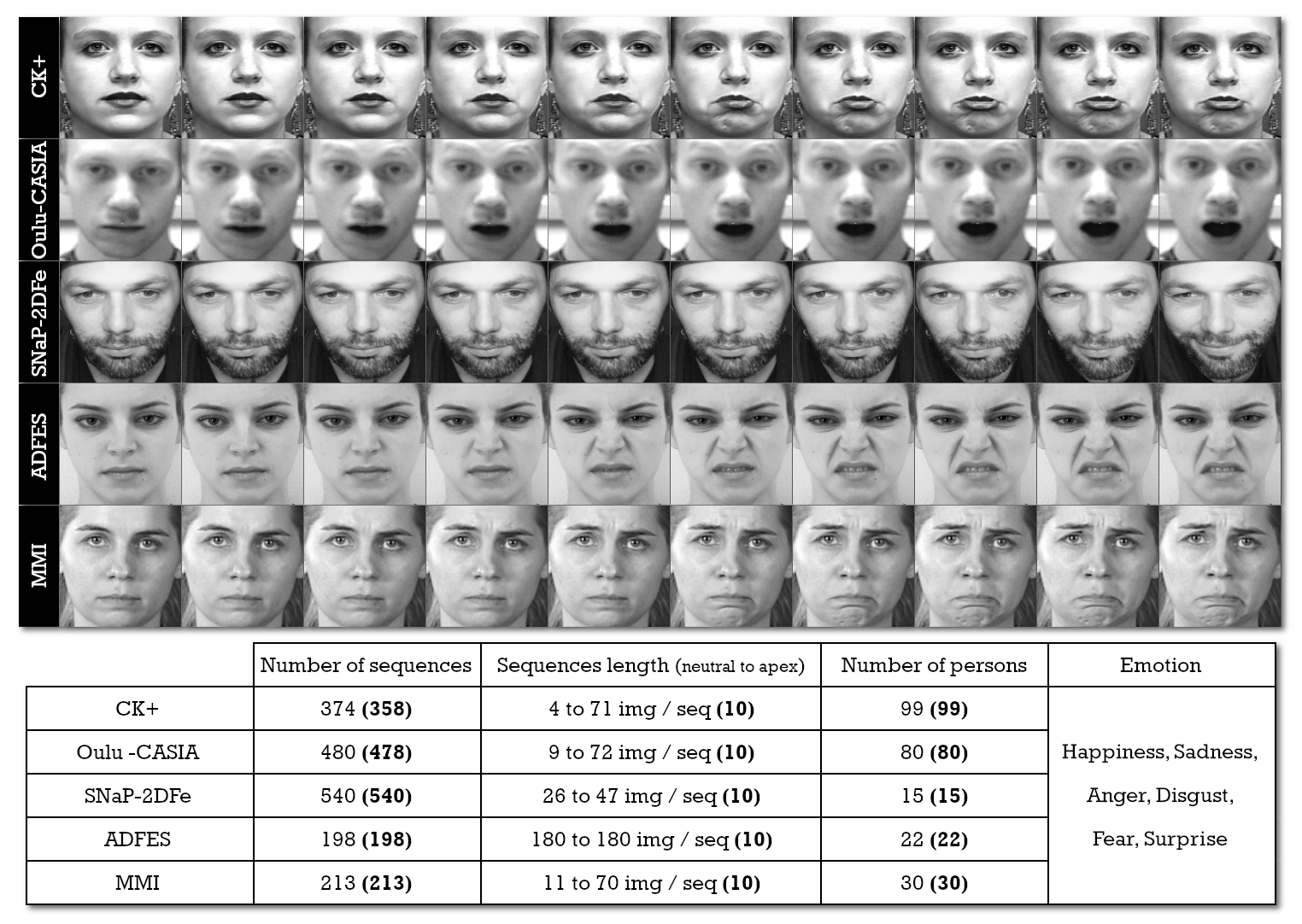}
\caption{\benjamin{Datasets used to analyze facial expressions from optical flow. The information in bold represents the final data obtained after the standardization process.}}
\label{fig:bdd}
\end{figure}

We hence combine several datasets, specifically the CK+ \cite{lucey2010extended}, Oulu-CASIA \cite{zhao2011facial}, \benjamin{ MMI \cite{pantic2005web}, ADFES \cite{van2011moving},} and SNaP-2DFe \cite{allaert2018impact} datasets, which contain the six basic expressions (anger, disgust, fear, happiness, sadness, and surprise). A brief overview of each dataset is provided here for completeness:

\textbf{CK+} contains 593 acted facial expression sequences from 123 participants, with seven basic expressions (anger, contempt, disgust, fear, happiness, sadness, and surprise). In this dataset, the expression sequences start in the neutral state and finish at the apex state. As illustrated in Figure \ref{fig:bdd}, expression recognition is completed in excellent conditions, because the deformations induced by the ambient noise, facial alignment and intra-face occlusions are not significant with regard to the deformations that are directly related to the expression. However, the temporal activation pattern is variable in this dataset, and spreads from 4 images to 66 images with a mean sequence length of $17.8\pm7.42$ images.

\textbf{Oulu-CASIA} includes 480 sequences of 80 subjects taken under three different lighting conditions: strong, weak and dark illuminations. They are labeled with one of the six basic emotion labels (anger, disgust, fear, happiness, sadness, and surprise). Each sequence begins in the neutral facial expression state and ends in the apex state. Expressions are simultaneously captured in visible light and near infrared.

\benjamin{\textbf{MMI} contains 213 sequences from 30 participants instructed to reproduce the six universal expressions (anger, disgust, fear, happiness, sadness, and surprise). Compared to the CK+ and the Oulu-CASIA datasets, this training dataset provides data closer to natural conditions, where participants are free to move their \marius{head} and their expressions are more spontaneous. These data tend to \marius{challenge} 
the robustness of the systems in less controlled acquisition conditions.}

\benjamin{\textbf{ADFES} include 198 sequences from 22 subjects \marius{(10 females and 12 males)}, out of which 10 are Mediterranean and 12 are North-European. 
From the nine recorded emotional states, we selected the six basic popular emotions (anger, disgust, fear, happiness, sadness and surprise) for our experiments.}

\textbf{SNap-2DFe} contains 1260 sequences of 15 subjects eliciting various facial expressions. These videos contain synchronized image sequences of faces in frontal and in non-frontal situations. For each subject, six head pose variations combined with seven expressions were recorded by two cameras, which results in a total of 630 constrained recordings captured with a helmet camera (i.e., without head movement) and 630 unconstrained recordings captured with a regular camera placed in front of the user (i.e., with head movements).

Concerning the above, we are using a subset of CK+ containing 374 sequences (which are commonly used in the literature). We use this subset to evaluate the `six universal expressions recognition problem'. For SNaP-2DFe, we only use the subset acquired by the helmet camera, used to remove head pose variations. All faces from the different databases are rotated and cropped (based on 68 landmark locations), color normalized \cite{coltuc2006exact} and resized in order to standardize the data, as illustrated in Figure \ref{fig:bdd}.

\subsection{Temporal standardization process}

As the duration of the different sequences across the datasets varies widely (from 4 to \benjamin{180} images), a temporal normalization of the sequences is necessary. Two temporal normalizations have been applied: TIM2, where the optical flow is calculated only between the first image (neutral) and the last image --- where the intensity of expression is at its highest (apex) --- and  TIM10, where 10 images are selected from the first image (neutral) to the last image (apex). For practical reasons, only sequences with at least 10 images are used in the evaluations, both for TIM2 and TIM10. The TIM2 case study analyzes the capacity of optical flow approaches to characterize facial expressions with high amplitudes that generally induce large discontinuities of movement (large displacements). As for the TIM10 case study, it provides an analysis on the capacity of optical flow approaches to maintain coherence of movement during the progressive activation of facial expressions.

\begin{figure}[!h]
\centering
\includegraphics[width=0.95\columnwidth]{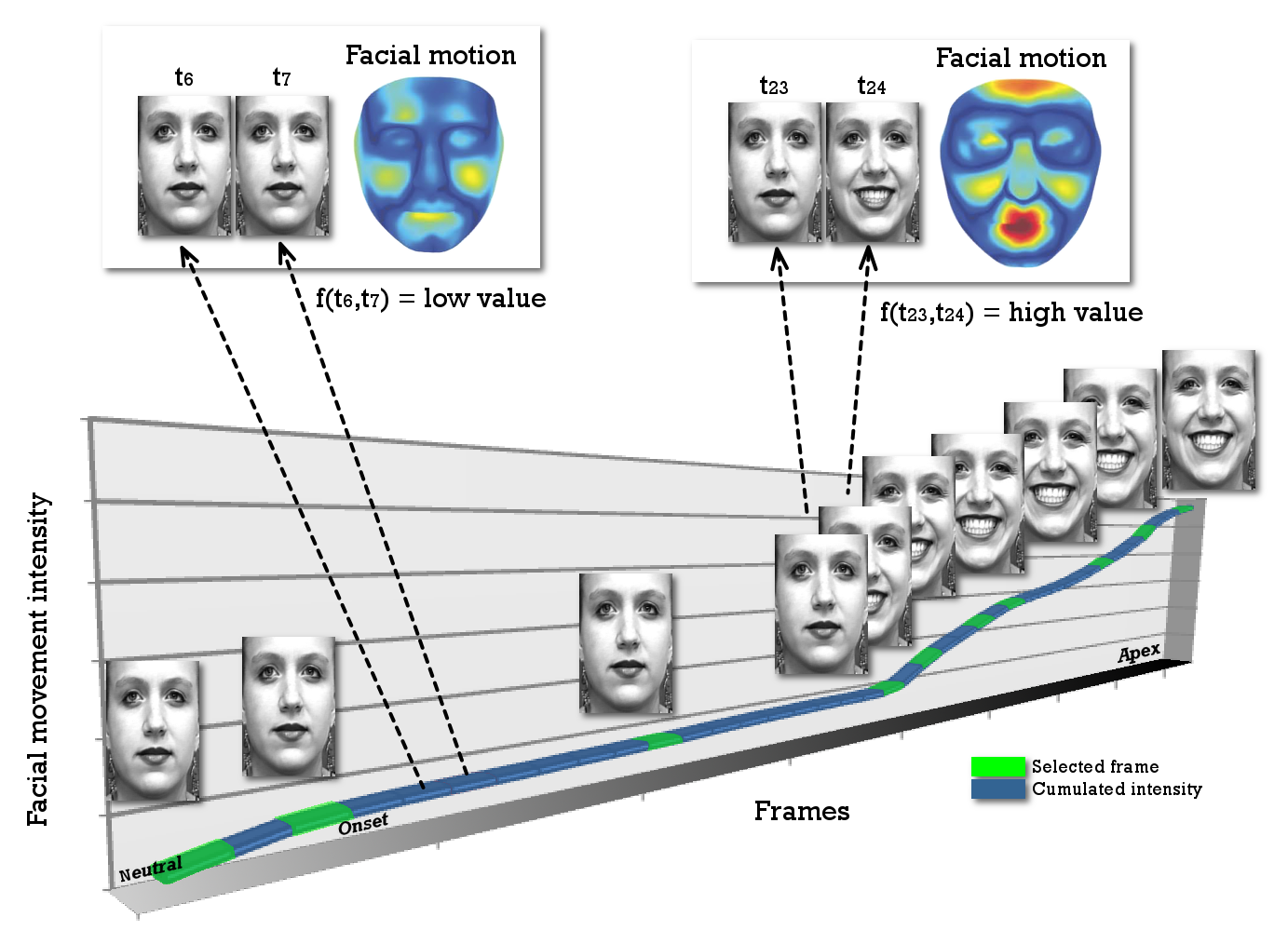}
\caption{Selection process of the key images according to the intra-face motion.}
\label{fig:curve}
\end{figure}

In the TIM10 case, in order to select the key images within a sequence, we calculated the intra-face motion intensity induced by the expressions. To avoid considering images where head movement is more pronounced than information relating to the facial expression, we have dissociated the movement from the rigid parts of the face (contour, nose) and from the dynamic facial elements (eyebrow, eyes, mouth). The movement between two successive images can be calculated as follows:

\begin{equation}
  f(t_{1},t_{2})=\begin{cases}
     \frac{\Delta_{E} + \Delta_{M}}{\Delta_{H}} , & \text{if $\Delta_{H}\geq0$ and $\Delta_{E}$ + $\Delta_{M}>0$}.\\
    0, & \text{otherwise}.
  \end{cases}
\end{equation}

\noindent where $\Delta_{E}$, and $\Delta_{M}$ represent the motion intensity in the dynamic regions of the face (eyebrows, eyes, mouth) between the two images $t_{1}$ and $t_{2}$, while $\Delta_{H}$ represents the motion intensity in the rigid regions of the face (nose, contour). The intensity of the rigid regions corresponds to the ratio of white pixels calculated in these regions. The more the head moves, the more the ratio is important.  Following this rule, if the face is not affected by any variations ($\Delta_{H}$ = $\Delta_{E}$ = $\Delta_{M}$ = 0) or if the head movement is too large ($\Delta_{H} > \Delta_{E} + \Delta_{M}$), the value obtained will be low, implying that the associated image is not significant. An illustration of the key images selection process is shown in Figure \ref{fig:curve}. We select the $n$ key images where the delta has changed the most between two successive images during the sequence (corresponding to the green segments in Figure \ref{fig:curve}).

\subsection{Performance criteria}
\label{perfcriteria}

The optical flow approaches are each evaluated using SVMs of type C-SVC with linear kernels (with C=1 and weight=1). We are aware that SVMs may not provide the highest classification accuracy, and that the reported results could be further optimized. However, our goal here is to compare optical flow approaches against a common benchmark. Hence, in the following evaluations, we focus primarily on the behavior of the different optical flow approaches and not on optimizing their performance for facial expression recognition.

To evaluate the performance of optical flow approaches for facial expression analysis, we use a 60-40 train/test validation protocol. The performance criteria being considered in our results are formulated as follows:

\textbf{AUC.} When using normalized units, the area under the curve is equal to the probability that a classifier will rank a randomly chosen positive instance higher than a randomly chosen negative one (assuming 'positive' ranks higher than 'negative'). For a predictor \textit{g}, an unbiased estimator of its AUC can be expressed by :

\begin{equation}
AUC(g) = \frac{\sum{}_{t_{0} \in D^{0}} \sum{}_{t_{1} \in D^{1}} 1[g(t_{0}) < g(t_{1})]}{|D^{0}| \cdot |D^{1}|}.
\label{eq3}
\end{equation}

\noindent where, $1[g(t_{0}) < g(t_{1})]$ denotes an indicator function which returns 1 if $g(t_{0}) < g(t_{1})$ otherwise return 0; $D^{0}$ is the set of negative examples, and $D^{1}$ is the set of positive examples.

\textbf{Mean AUC.} In order to uniformly evaluate all optical flow approaches, we have randomly generated ten learning configurations. For each evaluation, we report the average of the AUC obtained on the different learning configurations calculated by the following equation:

\begin{equation}
\overline{AUC} = \frac{\sum {AUC(g)_{i=1}^{c}}}{c}.
\label{eq2}
\end{equation}

\noindent where $c$ is the number of learning configurations (c = 10).

In the case of the \textit{data augmentation} experiments (Section \ref{fusion}), performance is calculated on exactly the same ten 60-40 train/test validation configurations, in order to ensure uniform evaluation of all optical flow approaches in the presence of and absence of data augmentation. For each evaluation, we select one optical flow approach and augment the training data with the remaining optical flow approaches --- making sure not to take any data from the test set. We then calculate the \textbf{average accuracy} obtained for each of the configurations. 

\benjamin{\textbf{Overall scoring.} For each evaluation, we compute a ranking of the different optical flow approaches based on the performance obtained on all datasets. For a datasets, an average is calculated from the accuracy obtained for each descriptor and for each optical flow approach. The averages obtained on the different datasets are used to calculate a rank for each optical flow method. The sum of these ranks gives the overall score of an optical flow approach over the whole \marius{settings.}} 

\section{Evaluation of optical flow approaches}
\label{solo}

\begin{figure}[!h]
\centering
\includegraphics[width=\columnwidth]{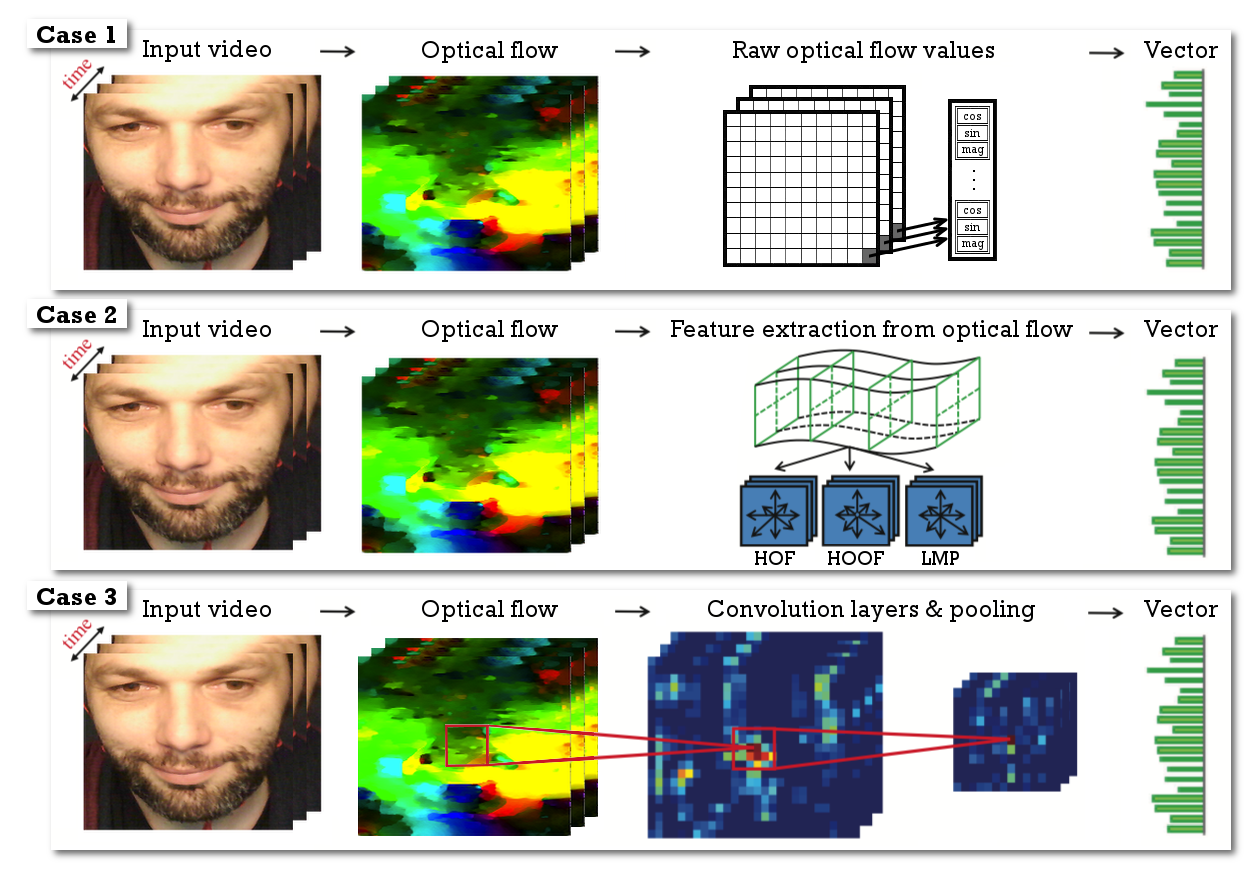}
\caption{Comparison of analysis from raw data, handcrafted and deep learning processes (based on optical flow and used for facial expression recognition).}
\label{fig:approach}
\end{figure}

The evaluation of optical flow approaches involves the application of the standard analysis process passing through the following steps: flow estimation, expression characterization and classification. In order to avoid any bias which may suggest that one analysis system is more suitable for one optical flow approach than another, three approaches  are investigated, as illustrated in Figure \ref{fig:approach}. They are as follows:

\begin{enumerate}
\item Analysis of the \textbf{raw} flow data input directly into the classifier;
\item Use of \textbf{handcrafted descriptors} to build a characteristic motion vector which is then passed to a classifier;
\item Use of \textbf{deep learning architectures} which rely on learned features constructed from the available data.
\end{enumerate}

\subsection{Analysis of the raw flow data}
\label{svm}

In this experiment, we directly evaluate the raw flow data obtained from the different optical flow approaches. This makes it possible to verify an optical flow approach's ability to preserve facial movements without using any descriptor or encoding. For this purpose, we use a linear SVM classifier. The use of this basic classifier makes it possible to avoid more complex learning approaches that could favour a particular optical flow approach.

\begin{figure}[!h]
\centering
\includegraphics[width=\columnwidth]{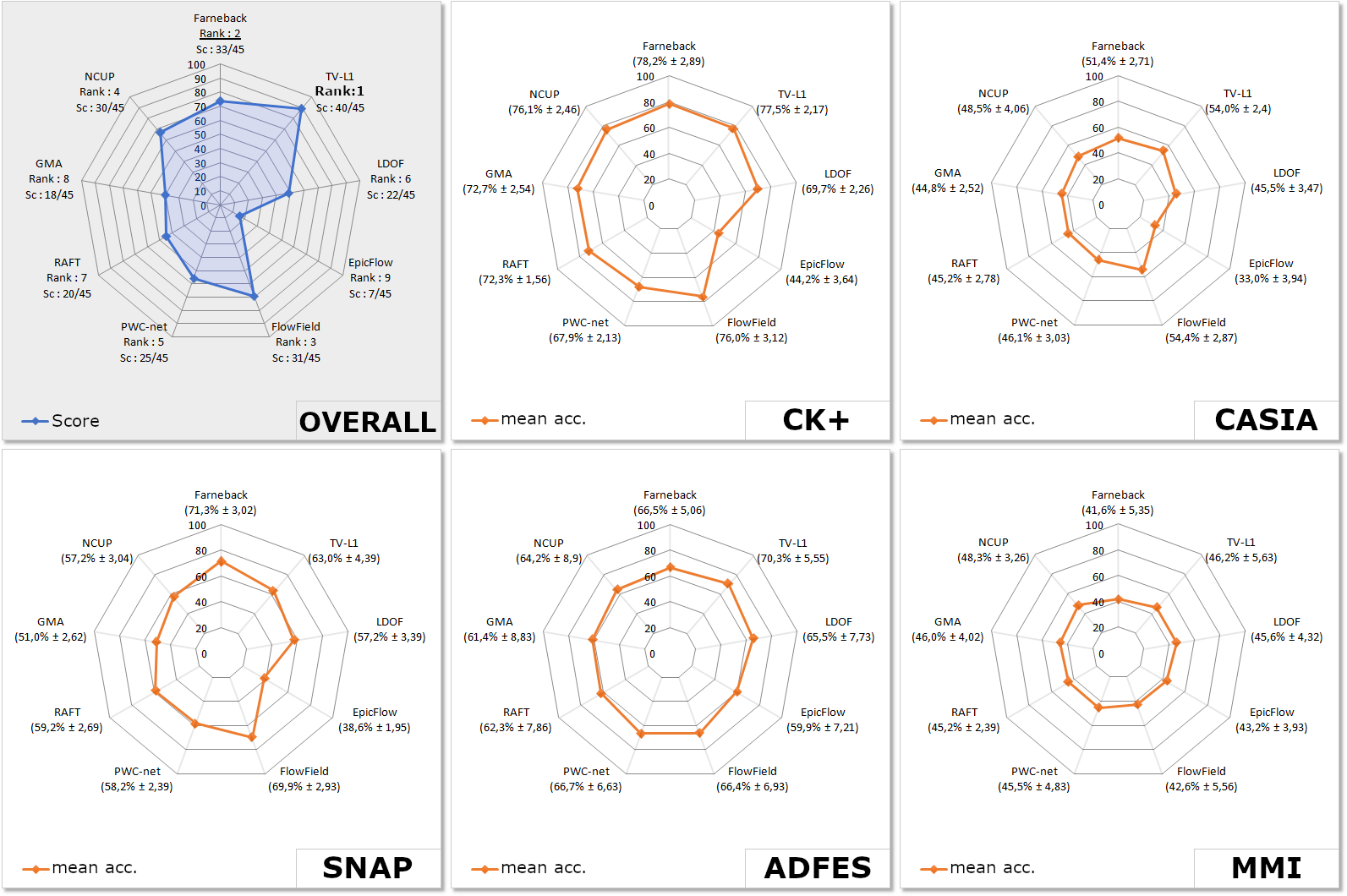}
\caption{\benjamin{Mean AUC obtained from analysis of the raw flow data with TIM2.}}
\label{fig:svm}
\end{figure}

In this experiment, we use only the standardized sequences with TIM2 (optical flow computed between two images: neutral and apex). The values representing the characteristic vector correspond to the raw optical flow values. Each pixel is characterized by two values: one for the direction, and one value for the magnitude of the motion. Since the images have a size of $50\times50$, the characteristic vector reaches a size of $50\times50\times2$ = 5000. Figure \ref{fig:svm} shows the results obtained from the various evaluations. 

The results obtained by applying a basic classifier suggest that there is a difference in performance between the different optical flow approaches. It is important to note that Farnebäck and TV-L1 methods that do not rely on recent motion approximation methods tend to provide better performance than recent approaches that perform well on MPI-Sintel such as EpicFlow based on a Deep Matching approach or PWC-net which are based on deep architectures. The performance of the Epicflow approach is relatively poor in comparison, largely because this approach is not well adapted to calculate the movement between two relatively different (distant in time) images, thus in the presence of important facial movements, mainly due to the matching method used. \benjamin{It is interesting to note that recent approaches such as RAFT, GMA and NCUP do not obtain the best performances despite their good performance on the MPI-Sintel \marius{dataset}. However, these methods stand out from the others on more complex bases such as MMI, where despite the normalisation of the faces, small displacements of the face occur.} Although the FlowField approach is less effective than PWC-net on datasets such as MPI-Sintel, it stands out from other recent approaches in the context of facial expression recognition and achieves competitive results in this problem domain. This is because, unlike other recent approaches, FlowField does not require explicit regularization, smoothing (like median filtering) or a new data term. Instead it solely relies on patch matching techniques and a novel multi-scale matching strategy which appears to be better adapted for characterizing facial movement. 

\subsection{Recognition from handcrafted approaches}

Most facial expression recognition systems use motion descriptors to more accurately characterize facial movements within the optical flow, to facilitate the classification step. To compare the performance of optical flow approaches using handcrafted approaches, and to avoid the possible bias that some descriptors might cause on a specific optical flow approach, we use several motion descriptors that are currently used in the area of facial expression recognition: HOF \cite{essa1997coding}, HOOF \cite{chaudhry2009histograms} and LMP \cite{allaert2018advanced}. All these descriptors are associated with a facial segmentation model in order to characterize the global facial movement. Among the existing models, we select a classic $5\times5$ grid in order to avoid any bias due to an incorrect estimation of the facial regions. As a reminder, in this evaluation, we do not seek to optimize the performance of the different approaches, only to propose a fair comparison between them. 

\begin{figure}[!h]
\centering
\includegraphics[width=\columnwidth]{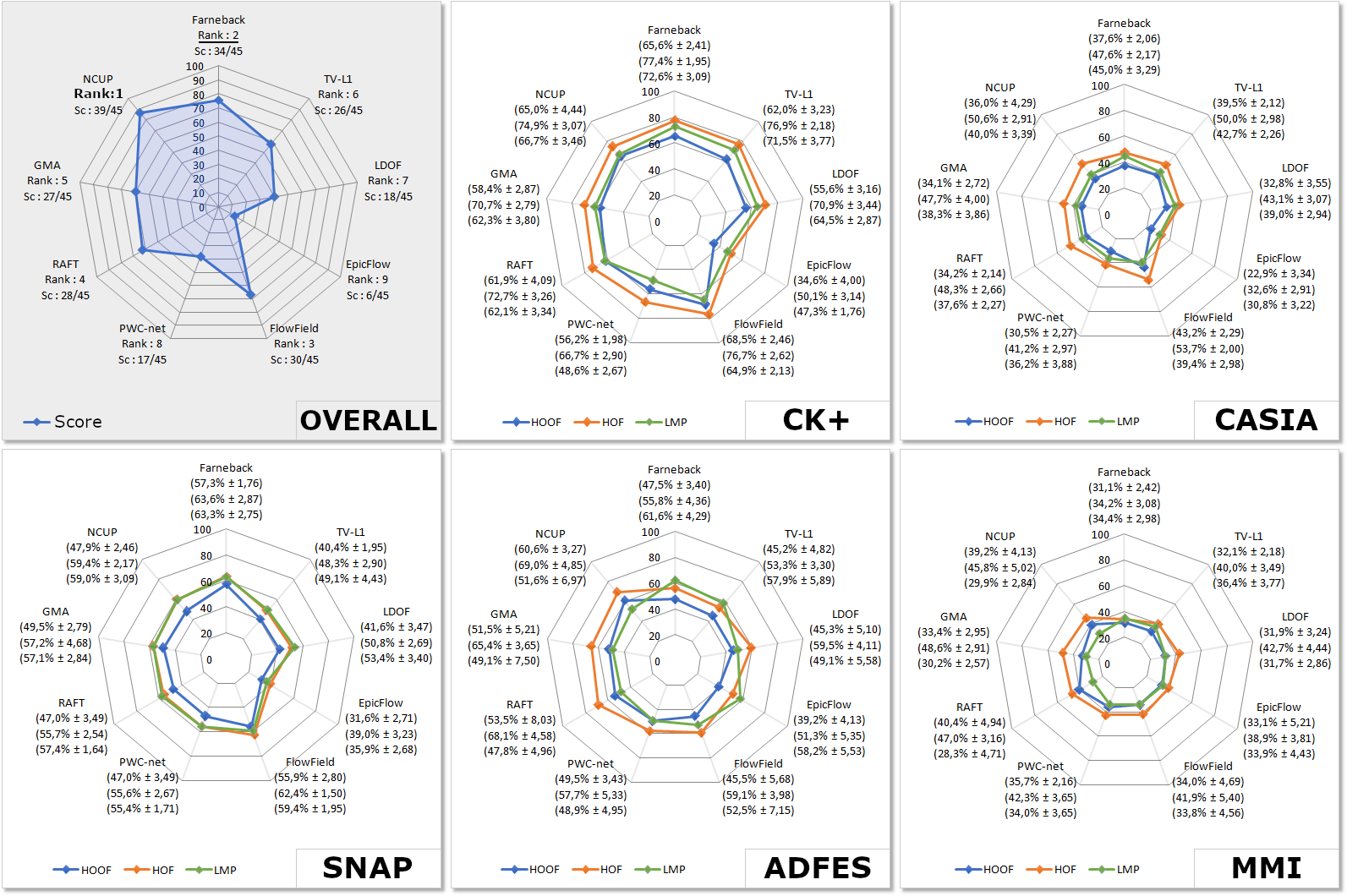}
\caption{\benjamin{Mean AUC obtained from the handcrafted approaches with TIM2.}}
\label{fig:hand_1}
\end{figure}

\begin{figure}[!h]
\centering
\includegraphics[width=\columnwidth]{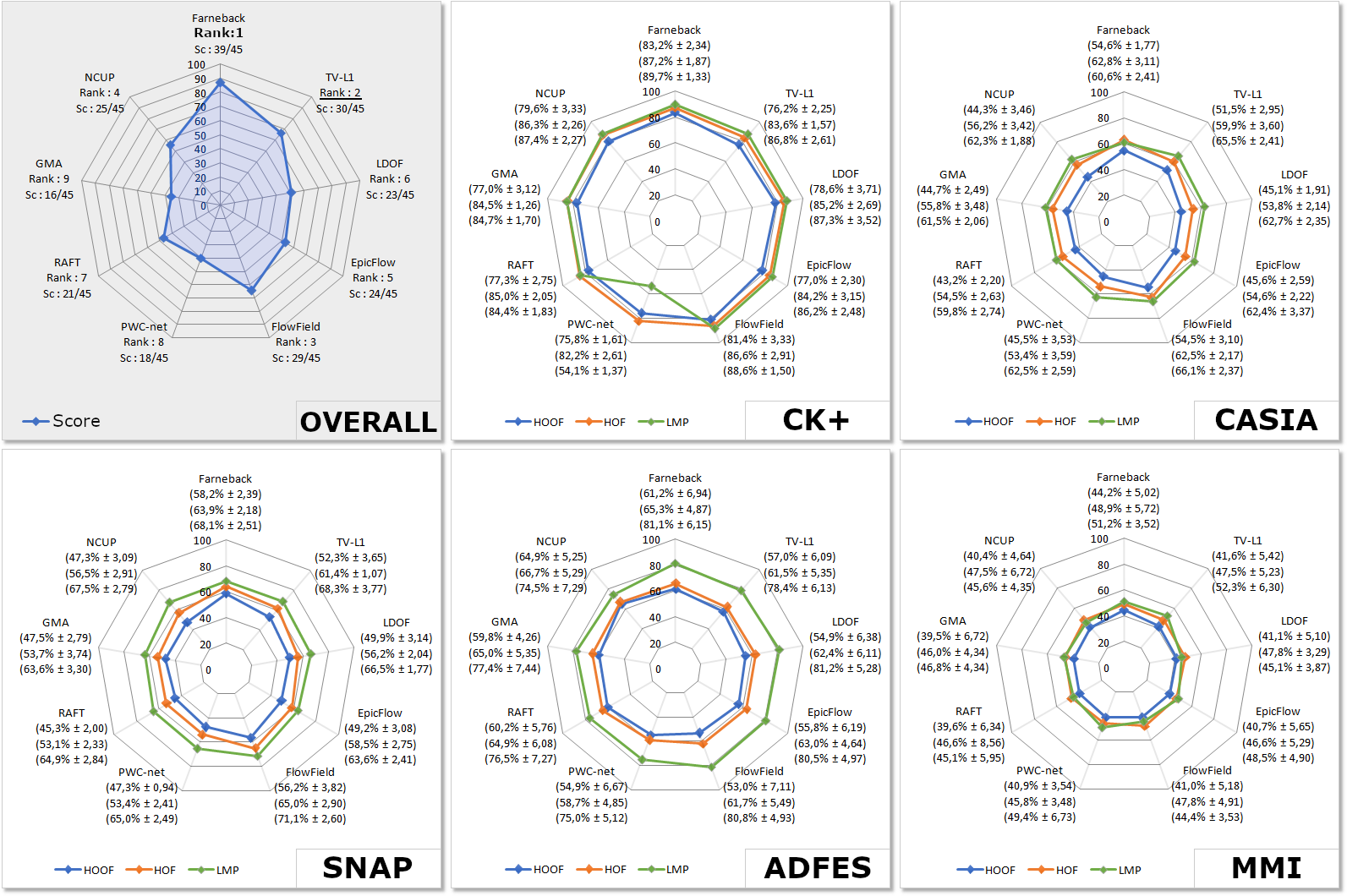}
\caption{\benjamin{Mean AUC obtained from the handcrafted approaches with TIM10.}}
\label{fig:hand_2}
\end{figure}

Figure \ref{fig:hand_1} shows the results obtained from the evaluations of different descriptors with the TIM2 configuration (motion between the neutral and the apex image) and Figure \ref{fig:hand_2} with the TIM10 configuration (which takes into consideration the movement throughout the activation sequence). To account for the movement, we calculate the characteristic vector within 25 regions of the face using the descriptor. Then, we construct a temporal vector by summing the different characteristic vectors. For all the descriptors, we analyze the distribution of the local movement over 12 directions. The characteristic vector reaches a size of $12\times25$ = 300.

Based on the results obtained from Figures \ref{fig:hand_1} and \ref{fig:hand_2}, two optical flow approaches repeatedly achieve very good performances: Farnebäck and FlowField. The difference in performance on the five datasets is explained by the fact that face registration is more complex on ADFES, CASIA and MMI and generates more residual noise which is reflected in the optical flow. In addition, the movement patterns of the expressions are more varied (e.g., intensity, direction) which makes the classification task more complex.

In Figure \ref{fig:hand_1}, where we do not consider temporal information, NCUP and Farnebäck outperform almost all the other approaches regardless of the descriptor used, closely followed by the Flowfield approach. As in the previous evaluation, the Epicflow approach gives the worst performance because it is not adapted to encode significant movements between two images, due to the matching method used. \benjamin{As earlier, the recent GMA, RAFT approaches perform best on the more complex \marius{datasets} such as ADFES and MMI. However, they do not compete with older approaches on more controlled datasets. Although NCUP is based on RAFT and PWCnet methods, it is found to perform very well in a TIM2 configuration. This highlights the contribution of upsampling approaches within the coarse-to-fine optical flow CNN, in an end-to-end fashion to allow optical flow networks to exploit fine details during learning.}

In Figure \ref{fig:hand_2}, it is observed that taking into account temporal information (TIM10) provides a better characterization of facial expressions. In this context, we observe that the Farnebäck and TV-L1 approaches remains very competitive with the FlowField approach regardless of the descriptor used. These results show that the two approaches tend to provide more consistent movements over time than the other studied approaches. Although the performance of the Epicflow approach is always lower, it can be seen that the performance is relatively similar to the performance of the other approaches. This is because the distance between the images is less important, and the movement at the pixel level is more coherently encoded. \benjamin{Although the results are better in the TIM10 configuration, the RAFT, GMA and NCUP approaches are ranked \marius{lower}. This can be explained by the fact that the accumulated optical flows on the different frames contain less consistent information, which reduces the performance of the classifiers.}

\subsection{Recognition from using deep-learning based approaches}

\begin{figure}[!h]
\centering
\includegraphics[width=\columnwidth]{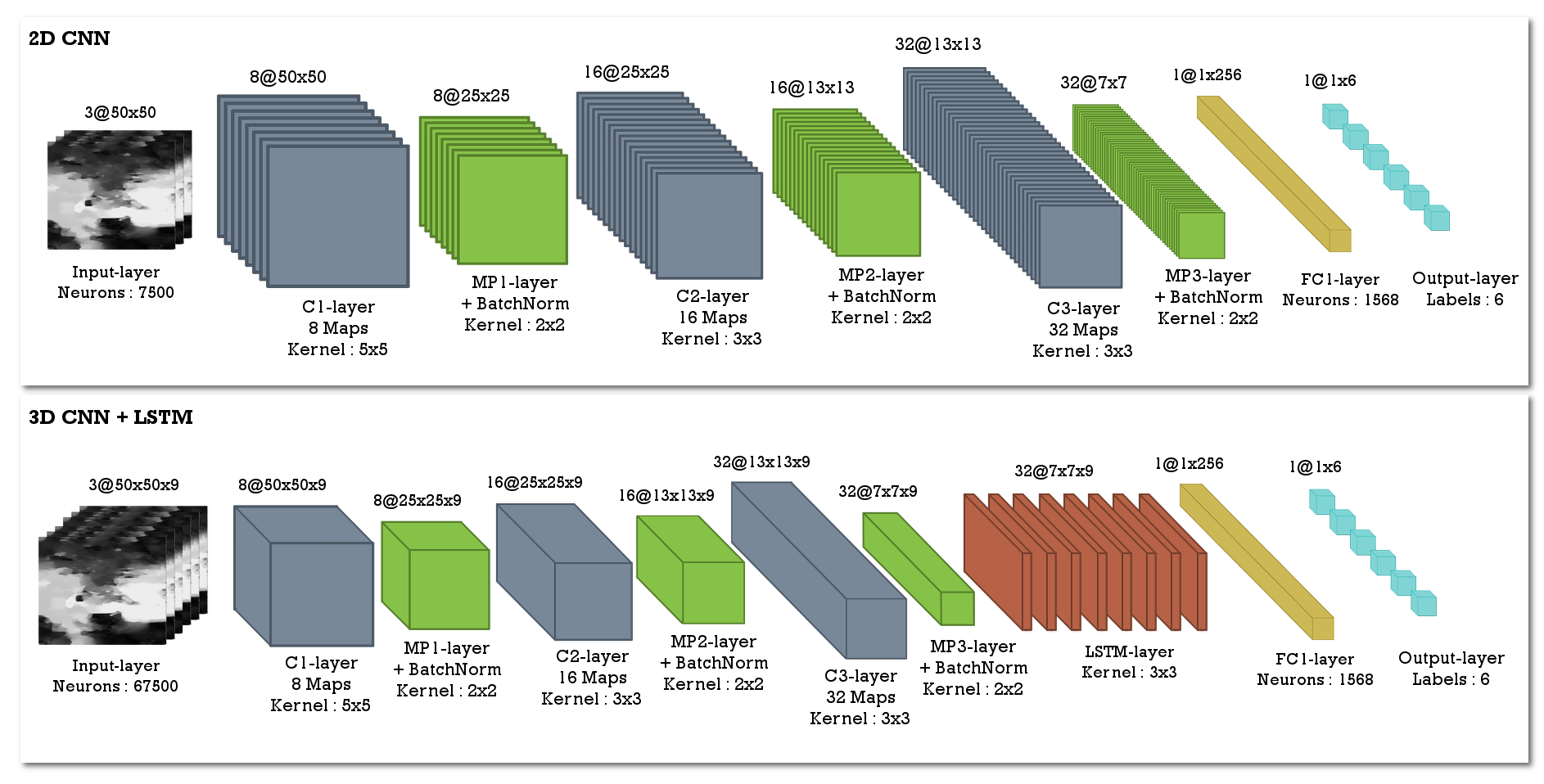}
\caption{Neural architectures used in the evaluations (C : Convolutional layer, MP : Max pooling, FC : Fully-connected layer).}
\label{fig:neural}
\end{figure}

In this experiment, we compare the results of different deep learning architectures when applied to different optical flows. In order that this study case can be properly compared to other studies cases, we choose to passing the inferred optical flow into a CNN, instead of learning the spatiotemporal representation from 3D convolution. Among the deep learning architectures used in computer vision \cite{khan2018guide}, we have selected two main types of architectures: Convolutional Neural Networks (CNNs) (based on the optical flow computed from the neutral and the apex image) and Recurrent Neural Networks (RNNs) which take into account the temporal information (all images in the sequence from the neutral to the apex image). The two architectures that are used in this evaluation are shown in Figure \ref{fig:neural}. Each architecture is applied to the different datasets and optical flows. 

\begin{figure}[!h]
\centering
\includegraphics[width=\columnwidth]{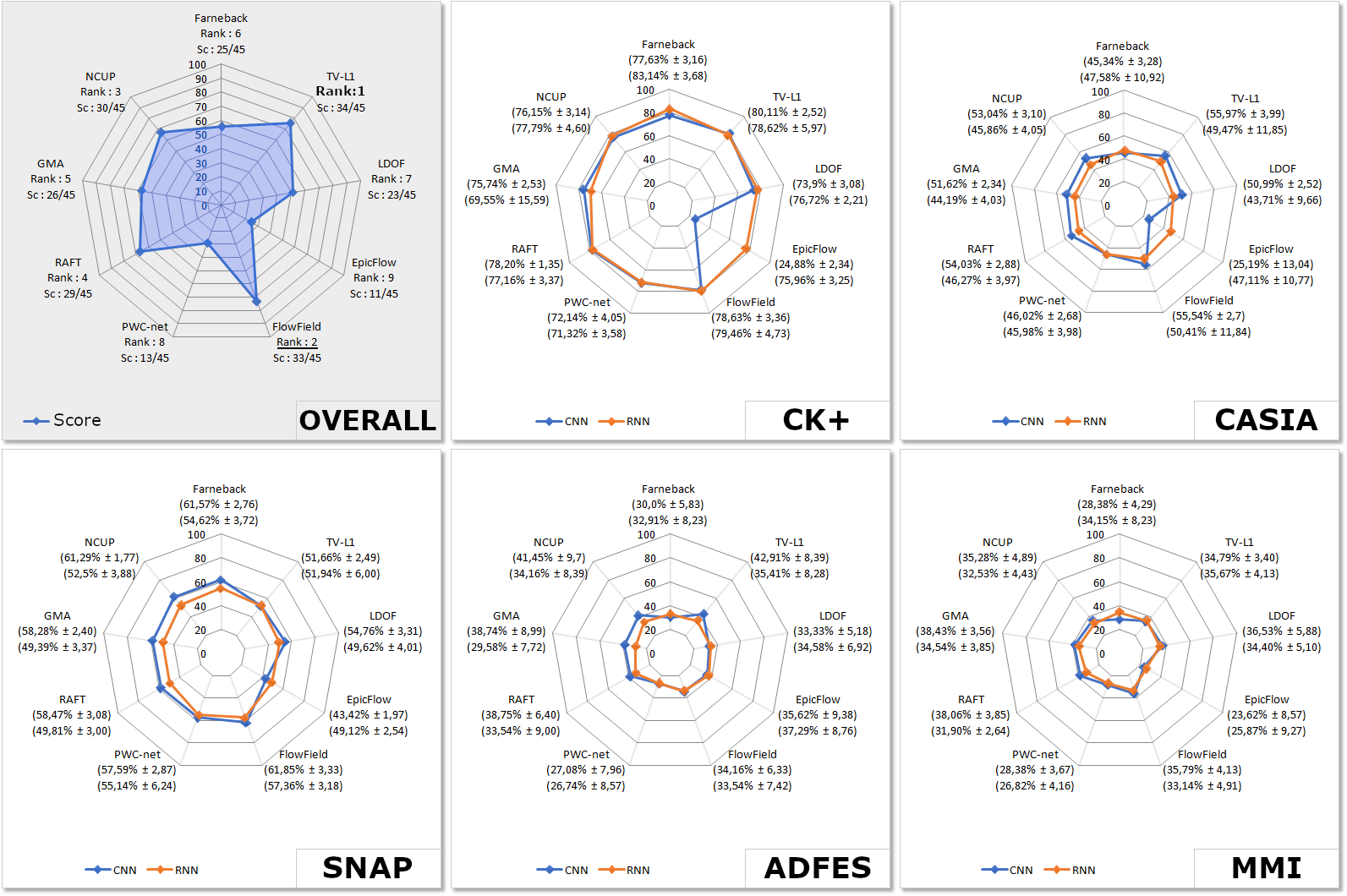}
\caption{\benjamin{Mean AUC obtained from the learning based approaches with TIM2 and TIM10.}}
\label{fig:deep}
\end{figure}

We are aware that there are other more complex architectures which produce a much better performance. However, in this evaluation, we simply intend to compare the different optical flow approaches and consider how they perform in low complexity contexts (to minimize learning biases). For the learning data, we use the same data format as the one used in the evaluation in Section \ref{svm}. Each motion pixel is characterized by two values: direction and magnitude. Since the images have a size of $50\times50$, the characteristic vector reaches a size of $50\times50\times2$ = 5000. For all evaluations, we use a batch size of 8 and an 10 epochs for training. Figure \ref{fig:deep} shows the results obtained for the various evaluations.  

Considering the results given in Figure \ref{fig:deep}, the performances of the different optical flow approaches are similar for both deep learning architectures (i.e., CNNs and RNNs). The performance of the optical flow approaches is relatively similar to the performances observed in previous evaluations. The TV-L1 and FlowField methods give the best performance using both CNNs and RNNs. Once again, the Epicflow and PWC-net approaches give the worst performance. In view of the performances obtained, it can be concluded that the strategy of these two approaches to propagate movement in the neighboring regions seems poorly adapted to the filter noise and may induce in turn noisy facial movements. \benjamin{The poor performance observed on the ADFES and MMI databases \marius{might be} 
related to the number of \marius{available training} data, which is not high enough to fully exploit the potential of the deep approaches. A more in-depth study on data augmentation is carried out in Section \ref{fusion} and validates this hypothesis.}

\subsection{Discussion of the optical flow evaluations}

Each of these evaluations highlight the significance behind the choice of the optical flow approach for facial movement analysis --- an incorrect choice can result in a significantly poorer performance. To fairly compare the different optical flow approaches, all approaches have been analyzed under the same conditions, ensuring that any bias that could result from the classifier optimization or the model selection has been omitted. 

In these evaluations, we selected different optical flow approaches which each have their own specific characteristics (See Section \ref{ssec:background}). 

\begin{figure}[!h]
\centering
\includegraphics[width=\columnwidth]{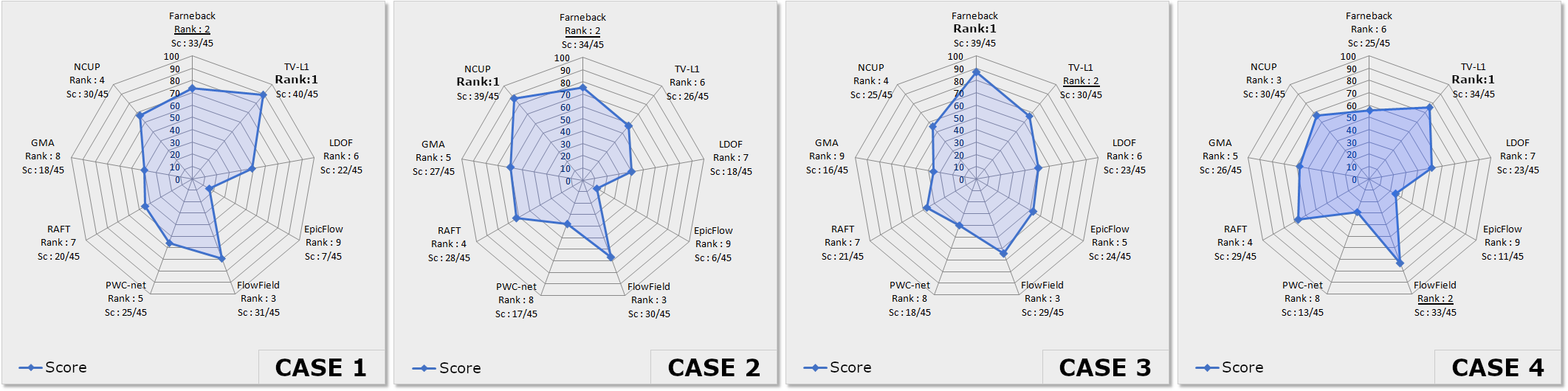}
\caption{\benjamin{Ranking of optical flow approaches according to the averages obtained per case study: case 1 (raw data TIM2 + SVM), case 2 (handcrafted descriptor TIM2 + SVM), case 3 (handcrafted descriptor TIM10 + SVM) and case 4 (CNN TIM2 and RNN TIM10).}}
\label{fig:recap}
\end{figure}

\benjamin{As highlight in Figure \ref{fig:recap}, the set of results obtained on the different evaluations makes it possible to distinguish four highly performant approaches among those which have been evaluated: Farnebäck, TV-L1, FlowField and NCUP. It is interesting to note that recent approaches such as RAFT, GMA and PWC-net, which have proven their effectiveness on optical flow benchmarks such as MPI-Sintel, seem less efficient \marius{in coping with} facial movement \marius{challenges}.} 

In view of the different studies carried out, the main reason for the difference in performance between the optical flow approaches analyzed seems to be the approximation of motion used to deal with large displacements and motion discontinuities. The intention behind the motion approximation techniques is to find the closest match visually. However, in the presence of important motion discontinuities and aperture problems, often present in the face context, the inferred movement is often different from the real movement. 

Since the EpicFlow approach demonstrated its performance on the MPI-Sintel dataset, many new approaches have the tendency to rely on these techniques to overcome movement discontinuities. However, when analysing facial movement, some discontinuities of movement can provide discerning information (e.g., wrinkles), which can be key in characterizing facial expressions. In this case, approaches based on approximation of motion techniques not adapted for facial expression movement tend to interpret these movements as noise because there is no local consistency in the propagation of motion. This highlights the paradox studied in this article. Indeed, the comparison criteria of optical flow approaches based on MPI-Sintel are not adequate, with the requirements expected from these approaches, for the analysis of facial expressions. Although the issues addressed in MPI-Sintel are identical to those observed in facial motion analysis, some constraints must be addressed in a different way, especially with respect to outlier filtering and motion approximation.

\benjamin{Regarding recent deep learning approaches (GMA, RAFT), they mitigate the complexity of the input data by producing flow predictions at a quarter of the resolution, which are \marius{upsampled} 
using bilinear interpolation during the test. However, this implies that fine details are usually lost and \marius{a post-processing step is required in order to restore them.} 
\marius{This post-processing can induce noise and affect the initial information}. The NCUP method differs from these approaches by modifying the upsampling method to reduce this bias.}

If we were to recommend an optical flow approach that would best characterize facial movement, we would choose either the Farnebäck, the FlowField or the NCUP approach (see Figure \ref{fig:intro}). The main advantage of the Farnebäck approach is that it is fast to calculate, which is an important feature to have if one wants to deploy a real-time analysis system. This can be combined with a good filtering algorithm, such as the one used by the LMP descriptor \cite{allaert2018advanced}. This filtering algorithm is based on the properties of facial movement propagation and can be used to improve performance. As for FlowField, it is based on a rather complex matching algorithm that is relatively more computationally expensive, especially when evaluating on a CPU. Still, FlowField has shown its effectiveness on the MPI-Sintel benchmark and on characterizing facial movement. \benjamin{NCUP takes into account the latest technical solutions in the literature. Although it does not achieve the best performance on all previous evaluations, it is the best in its category and suggests a perspective for future deep learning approaches.}

Now, it is important to consider the relevance of calculating a perfect optical flow that would be applicable to all problems. With the large number of optical flow approaches proposed in the literature, we explore the construction of a unique augmented model which relies on a set of the most common model characteristics.

\section{Data augmentation by optical flows}
\label{fusion}

Instead of identifying the most appropriate optical flow approaches to characterize facial movement, we study whether it is possible to rely on the properties of the different optical flow approaches in order to build a unique approach for analyzing facial expressions. With the capabilities of learning-based approaches, we explore in this section whether it is possible to use different combined optical flow approaches to artificially augment learning data. 

To assess the impact of data augmentation based on optical flow approaches, we use the CNN architecture in Figure \ref{fig:neural} with the TIM2 configuration on the three databases which were used in the earlier experiments (see Section~\ref{solo}). We choose the TIM2 configuration over the TIM10 configuration, as working on sequences is much more time-consuming and memory-intensive, especially if one wants to study a multitude of data augmentation methods. Additionally, if the augmentation provides better encoding of movement information between two images, it is expected that the results should improve when considering two successive images. To ensure that the contribution of the data augmentation is accurately compared, at the expense of the performance that can be achieved, we set all random parameters consistently: the random seeds are fixed at the initialization of the learning, the initial weights of the layers and the constant biases are the same for all runs, and the learning data is fixed according to the studied configurations. 

\subsection{Evaluation of the data augmentation approach}

\benjamin{In lieu of the performances obtained by the different optical flow approaches analyzed in the previous section, we decide to study the contribution of the data augmentation process on only three approaches: the Farnebäck, FlowField and NCUP approaches. These three approaches have been selected because they tend to provide good performances for characterizing facial movement (see Section \ref{solo}) and \marius{cover} the different categories of approaches proposed in the literature.}

\begin{figure}[!h]
\centering
\includegraphics[width=\columnwidth]{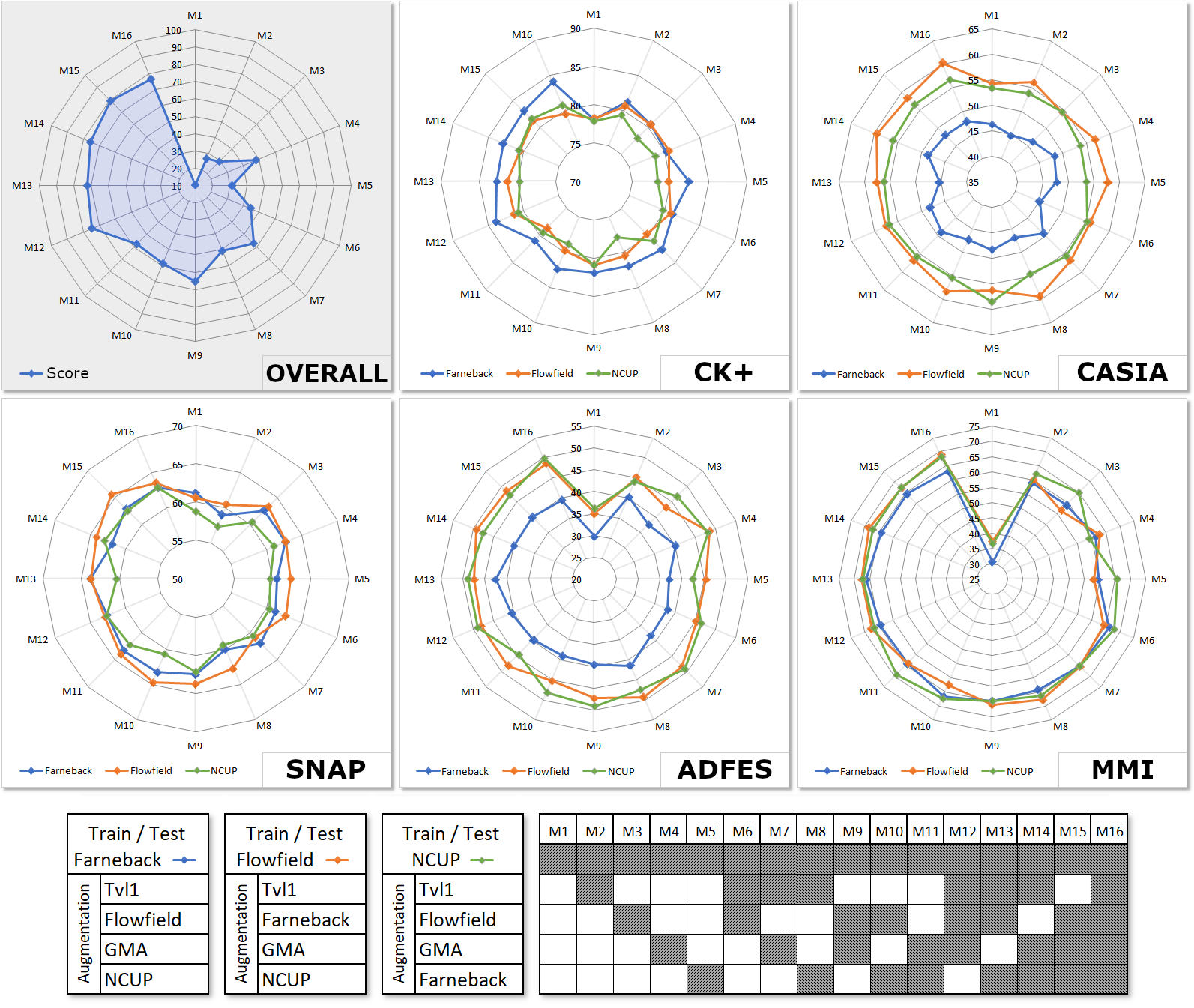}
\caption{\benjamin{Data augmentation based on optical flow. The tables in the bottom row represent the different augmentation configurations used for each optical flow method. The blackened boxes represent the optical flow approaches which are used for data augmentation.}}
\label{fig:aug}
\end{figure}

For each of the datasets, we iterate through all of the different possible data augmentation configurations. The results are shown in Figure \ref{fig:aug}. \benjamin{For each datasets, the \marius{reported} accuracy corresponds to the results obtained by taking an optical flow approach as a train/test and with or without other optical flow approaches for augmentation. The different blackened boxes in the table at the bottom of the figure represent the optical flow approaches which are used for data augmentation\marius{. Sixteen configurations ranging $M1$ to $M16$ where considered. $M1$ corresponds with to a setting where no augmentation is used. $M16$ corresponds to an augmentation considering the Tvl1, Flowfield, GMA and NCUP.} 
The first column represents the results obtained without data augmentation and the last column represents the results obtained when using a data augmentation method which uses all the studied optical flow approaches. The results for each of the configurations is computed for each datasets, and overall performances \marius{are reported as the score introduced in Section~\ref{perfcriteria}.}}

When considering the results obtained across all of the datasets, we can see that there is a significant improvement in the performance of the different optical flow approaches when the initial data are artificially augmented using other optical flow approaches. \benjamin{The Farneback approach gains on average 5\% on CK+, 3\% on CASIA, 2\% on SNAP, 13\% on MMI and 36\% on ADFES. As for the FlowField approach, it gains on average 4\% on CK+, 6\% on CASIA, 5\% on SNAP, 14\% on MMI and 31\% on ADFES. \marius{The NCUP approach} gains on average 5\% on CK+, 5\% on CASIA, 4\% on SNAP, 13\% on MMI and 33\% on ADFES. The significant gains observed on the MMI and ADFES databases are mainly due to the small amount of training data contained in these databases, which does not facilitate the convergence of the classifiers. The proposed augmentation method \marius{overcome this difficulty. } 
Overall, we notice that \marius{the} more optical flow approaches are used in the augmentation process, the more the performance tends to increase. \marius{Besides, combining various optical flows techniques in order to increase the quantity of the available training data, might also result in an increased robustness to motion discontinuities, illumination changes or motion intensity variations that are partially dealt with by the various optical flow techniques selected.}} 

\subsection{Discussion of the data augmentation approach}

In the previous section, we investigated whether artificial data augmentation by optical flow can improve the performance of neural networks. By studying the three approaches that we have identified to be the most suitable for analyzing facial movements (Farnebäck, FlowField and NCUP), and we can see that artificial data augmentation based on other optical flow approaches can significantly improve performance (from 2\% to 33\% depending on the configuration).

We think it is interesting to use fast computational optical flow approaches such as the Farnebäck approach to characterize facial movement, while relying on other optical flow approaches such as FlowField and NCUP to enhance learning and overcome the flaws of the less robust approaches. In the case of neural networks, it would be advisable to perform offline learning with an extended set of optical flow approaches, where the computation time can be relatively long. Then, use a fast but not very robust optical flow approach to extract facial movement in a real-time system.

\section{Conclusion}
\label{conclu}

In this work, the main contribution lies in the performance analysis of different optical flow approaches in characterizing facial expressions. Our experiments clearly show that two approaches generally outperform all the others: Farnebäck and FlowField. The Farnebäck approach has the advantage of being quick to compute, while the FlowField method has proven its effectiveness both on facial movement analysis and on more complex datasets such as MPI-Sintel.

\benjamin{Despite the NCUP approach tends to be close to the performance obtained by traditional methods,} we have shown that the recent dense optical flow approaches that obtain the best performance on MPI-Sintel are not always well suited for the analysis of facial expressions. Although the issues addressed in MPI-Sintel are identical to those addressed in facial expression analysis (large displacement, motion discontinuity, occlusions), the solutions implemented to address them are not always ideal for both case studies.

Indeed, in the case of MPI-Sintel, where the images are very large and correspond to a synthetic film, the methods which obtain the best performance are generally those that rely on the best motion approximation approach \benjamin{or the best upsampling approach} to reduce the gap with the ground-truth. It is more interesting in this case to cover the majority of the pixels even if the approximated values are sometimes outliers. However, in facial motion analysis, it is crucial to accurately approximate the motion because all motion discontinuities are sometimes correlated with the facial expression and provide important information to characterize the expression. Indeed, some discontinuities in facial movement related to the activation of facial muscles can provide discernible information (e.g., wrinkles), which may be essential for discriminating between facial expressions. \benjamin{It is important to note that the GMA method, designed to solve \marius{the motion discontinuity problems} induced by occultations, does not \marius{perform} 
well in \marius{current settings}.}

The method used to estimate movement, in regions where there are motion discontinuities within the face, explains why, in this study, there is a strong difference between the results obtained by approaches such as EpicFlow and PWC-net when used to analyze facial expressions. Theses approaches based on motion approximation techniques tend to interpret the movements induced by the activation of facial muscles as noise because there is no local coherence in the propagation of the movement. The intention behind the motion approximation based on the neighboring regions of the noised regions is to find the closest match visually, which is often not identical to the expected motion. An important difference is that motion in neighboring regions is known to be very noisy with respect to the shift of neighboring pixels, whereas the optical flow is generally locally smooth and sometimes abrupt. For facial expression analysis, these methods are based on too rigid motion approximation approaches that tend to propagate incoherent motion that is not correlated with facial expression. \benjamin{Regarding recent deep learning approaches that use downsampling and post-processing methods to mitigate the complexity of the input data, the \marius{gain in performance} is limited. In our experiments, we \marius{notice} 
that this solution is not optimal and tends to reduce performance. The solution proposed in NCUP, which consists in incorporating the upsampling step directly into the learning process, is a promising improvement that seems to be better adapted to facial analysis.}

For facial expression analysis, the approximation strategy proposed in the FlowField approach seems to be a very good compromise to analyze both facial motion and motion in a scene. The particularity of this methods that they do not require explicit regularization or smoothing (such as in median filtering), but are instead a pure data-oriented search strategy which only finds the most inliers, while effectively avoiding the outliers. It is also possible to use more classical methods such as the pyramidal approach proposed by Farneback. These approximation strategies are less accurate but generate less error for the analysis of facial movements and remain relatively quick to calculate.

We have thus illustrated through our experiments that some optical flow approaches differ strongly in their effectiveness in characterizing facial movements, and that it is not always easy to find a single unique solution that is both robust and fast. As such, second contribution of this work was to propose \textit{and} benchmark a data augmentation method which combines \textit{multiple} optical flow approaches. We have indeed shown that the artificial augmentation of a training set in this way can improve the classification accuracy. The results produced show that on average, increasing data based on optical flow approaches can improve performance by 2\% to 33\%, depending on the optical flow approaches used to test the data and the test dataset which is being used. This has potential applications in in-the-wild on-line analysis, where a noisy but fast optical flow can encode on the fly the data while relying on a complex offline learning process where more robust and time-consuming optical flow approaches are used for data augmentation.

In order to improve the robustness of facial optical flow, specific datasets should be released to the community. MPI-Sintel does not seem adequate for expression related challenges (variations of pose, expressions, occlusions) or challenges relating to facial motion characteristics (in the context of an expression, a discontinuity can be a source of information and does not always have to be corrected). New datasets such as SNaP-2DFE \cite{allaert2018impact} which record the facial motion both in the presence or in the absence of head movements are opening the way to specific facial-expression benchmarks, but more effort should be invested in such work.

Ultimately, we believe that future work should consider the following three aspects: (1) encoding plausible facial physical constraints when extracting optical flow data, (2) the design of temporal architectures capable of modeling the temporal activation of facial expressions and (3) exploring intra-optical and inter-optical flow augmentation techniques.


\bibliography{mybibfile}

\end{document}